\documentclass[sigconf]{acmart}

\usepackage{bm}  

\newcommand{\m}{\mathbf{m}}
\newcommand{\cov}{\mathbf{V}}
\newcommand{\covp}{\mathbf{V_p}}
\newcommand{\covc}{\mathbf{V_c}}
\newcommand{\tril}{\mathbf{L}}

\newcommand{\triu}{\mathbf{U}}
\newcommand{\x}{\mathbf{x}}
\newcommand{\q}{\mathbf{q}}
\newcommand{\rvec}{\mathbf{r}}
\newcommand{\ident}{\mathbf{I}}

\newcommand{\CHANGED}[1]{}
\newcommand{\NEW}[1]{#1}

\copyrightyear{2024} 
\acmYear{2024} 
\setcopyright{rightsretained} 
\acmConference[SIGGRAPH Conference Papers '24]{Special Interest Group on Computer Graphics and Interactive Techniques Conference Conference Papers '24}{July 27-August 1, 2024}{Denver, CO, USA}
\acmBooktitle{Special Interest Group on Computer Graphics and Interactive Techniques Conference Conference Papers '24 (SIGGRAPH Conference Papers '24), July 27-August 1, 2024, Denver, CO, USA}
\acmDOI{10.1145/3641519.3657502}
\acmISBN{979-8-4007-0525-0/24/07}

\begin{document}

\title{\NEW{N-Dimensional Gaussians for Fitting of High Dimensional Functions}}

\author{Stavros Diolatzis}
\email{stavros.diolatzis@intel.com}
\orcid{0000-0001-6051-372X}
\affiliation{%
  \institution{Intel Labs}
  \streetaddress{}
  \city{}
  \state{}
  \country{France}
  \postcode{}
}

\author{Tobias Zirr}
\email{tobias.zirr@intel.com}
\orcid{0009-0003-1091-3858}
\affiliation{%
  \institution{Intel Labs}
  \streetaddress{}
  \city{}
  \state{}
  \country{Germany}
  \postcode{}
}

\author{Alexander Kuznetsov}
\email{alexander.kuznetsov@intel.com}
\orcid{0009-0001-7084-3391}
\affiliation{%
  \institution{Intel Labs}
  \streetaddress{}
  \city{}
  \state{}
  \country{USA}
  \postcode{}
}

\author{Georgios Kopanas}
\email{georgios.kopanas@inria.fr}
\orcid{0009-0002-5829-2192}
\affiliation{%
  \institution{Inria, Université Côte d’Azur}
  \streetaddress{}
  \city{}
  \state{}
  \country{France}
  \postcode{}
}

\author{Anton Kaplanyan}
\email{anton.kaplanyan@intel.com}
\orcid{0000-0002-8376-6719}
\affiliation{%
  \institution{Intel Labs}
  \streetaddress{}
  \city{}
  \state{}
  \country{USA}
  \postcode{}
}

\renewcommand{\shortauthors}{Diolatzis et al.}

\begin{teaserfigure}
  \includegraphics[width=\textwidth]{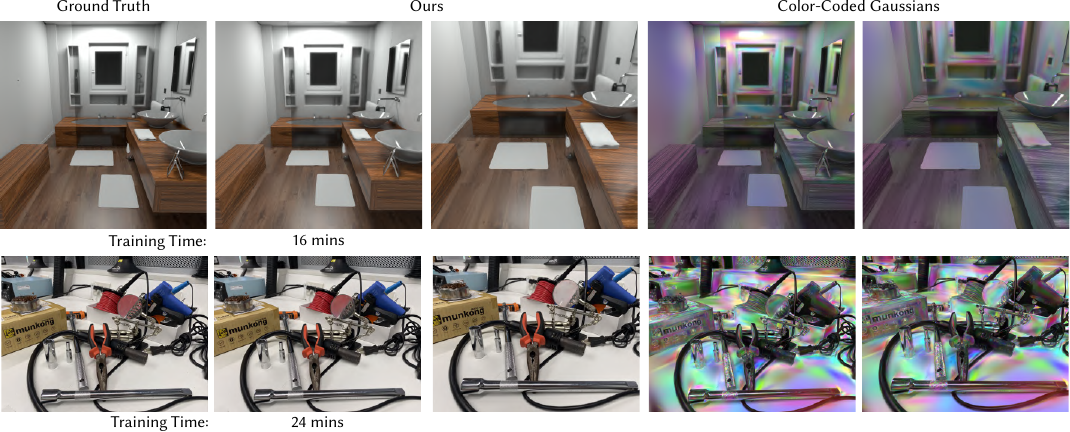}
  \caption{Our method optimizes N-Dimensional Gaussians to approximate high dimensional anisotropic functions in a few minutes. Our parameterization, culling and optimization-controlled refinement allows us to quickly estimate Gaussian parameters to represent surface (top) and volume radiance fields (bottom). Top: A synthetic scene with given geometry and material information creates a 10 dimensional input space. While the Gaussians live on surfaces \NEW{they can represent rough and specular reflections by moving faithfully with viewpoint. Bottom: Similarly, 6D Gaussians of position and direction correctly reconstruct the appearance through the magnifying glass while being consistent with camera movement}. We recommend watching the supplementary video to observe the movement of the complex effects.}
  \Description{Teaser figure.}
  \label{fig:teaser}
\end{teaserfigure}

\begin{abstract}
  In the wake of many new ML-inspired approaches for reconstructing and representing
  high-quality 3D content, recent hybrid and 
  explicitly learned representations exhibit promising performance and quality characteristics. 
  However, their scaling to higher dimensions
  is challenging, e.g. when accounting for dynamic content with respect
  to additional parameters such as material properties, illumination, or time.
  In this paper, we tackle these challenges for an explicit representations based on 
  Gaussian mixture models. With our solutions, we arrive at efficient fitting of
  compact N-dimensional Gaussian mixtures and enable efficient evaluation at render time:
  For fast fitting and evaluation, 
  we introduce a high-dimensional culling scheme 
  that efficiently bounds N-D Gaussians, inspired by Locality Sensitive Hashing.
  For adaptive refinement yet compact representation, we introduce a 
  loss-adaptive density control scheme that incrementally guides the use of 
  additional capacity towards missing details.
  With these tools we can for the first time represent complex appearance that 
  depends on many input dimensions beyond position or viewing angle within a compact, 
  explicit representation optimized in minutes and rendered in milliseconds.
\end{abstract}

\ccsdesc[300]{Computing methodologies~Reflectance modeling}
\ccsdesc[500]{Computing methodologies~Computer graphics}

\keywords{datasets, neural networks, gaze detection, text tagging}


\maketitle

\section{Introduction}

Computer graphics has recently seen an influx of new optimization-based approaches for reconstructing and representing high-quality graphics content such as complex geometry, lightfields, or high-fidelity avatars from multiview images. 
While inspired by machine learning, where neural networks are often
used to handle high-dimensional data, many
of these techniques are optimized for lower-dimensional domains
with well-understood frequency content (e.g., spatial or separable
spatio-angular domains).
In fact, purely ML-derived approaches, such as implicit neural
representations, where all information is implicitly stored in network weights,
are quickly outperformed by hybrid or
explicit representations. In these representations learned parameters are directly associated
with certain input dimensions of the overall represented distribution.
Thus, explicit disentanglement of certain query inputs such as
spatial coordinates can be used to reduce the data that needs to be accessed
per inference or query.
We observe that this trend towards more explicit representations has led to a limiting
of their usefulness for practical computer graphics problems that
require handling higher input dimensionality, e.g. to handle
high-quality lighting under changing conditions, paired with diverse material
properties.

In this paper, we recover this ability to scale to higher numbers
of input dimensions, for an explicit representation based on $N$-dimensional
Gaussian mixtures (GMMs). Efficient splatting of GMMs for lower-dimensional
radiance fields~\cite{kerbl20233d} recently revived interest in such
representations.
We evaluate two relevant applications, shading synthetic scenes with 
high appearance variability, and capturing scenes from images with
strong view-dependent effects, showcased in Figure~\ref{fig:teaser}. We tackle two particular challenges
in the construction of compact explicit $N$-dimensional Gaussian
mixtures: (1) recover training and evaluation efficiency for larger numbers
of mixture components, each with high-dimensional anisotropic distributions;
and (2) adaptive refinement of the mixture representation in the face of unknown 
domain-specific dependencies or inter-dimensional correlations.
We introduce:
\begin{itemize}
  \item An unconstrained $N$-dimensional adaptive Gaussian mixture representation,
  \item Fast fitting and evaluation based on a high-dimensional culling scheme for $N$-D Gaussians inspired by Locality Sensitive Hashing, and
  \item A domain-independent, loss-adaptive refinement scheme to arrive at compact and high-quality mixtures.
\end{itemize}

\noindent \NEW{Our code implementation is available at \href{https://github.com/intel/ngd-fitting}{https:\slash \slash github.com/intel/ngd-fitting}}.

\section{Related Work}

We review previous work related to high-dimensional fitting,
neural graphics, and rendering of learned representations.
We focus on the most relevant building blocks and refer to the survey
by Tewari et al.~\shortcite{tewari2022advances} for a more extensive overview. 

\subsection{High-Dimensional Anisotropic Distribution Fitting}

When fitting distributions varying with respect to high numbers of input parameters, 
correlations arising from mixed input types (e.g., spatial, angular, path-space, temporal) 
are hard to predict.
The problem of compactly representing high-dimensional anisotropic distributions 
is closely related to (unsupervised) 
clustering~\cite{figueiredo2002unsupervised}, where the meaning
of distance and similarity is reduced with increasing number of dimensions.
Locality Sensitive Hashing (LSH) techniques are an inspiration for our approach to practically bound and cull our mixture components.
The idea of projection onto random unit vectors~\cite{andoni2015practical} for localization and bounding proved particularly useful for us, as e.g. also used to cut down on attention computations for irrelevant tokens in Diffusion Models~\cite{kitaev2020reformer}.
To arrive at compact adaptive representations, many applications use domain knowledge about the structure of their high-dimensional data distributions,
in order to derive explicit clustering criteria, such as for high-energy paths 
in path space for path guiding~\cite{reibold2018selective}, or 
probes in radiance caching~\cite{krivanek2006making}.
In our general scenario, we build on careful iterative refinement strategies
such as commonly used for optimizing implicit neural representations~\cite{tewari2022advances} and explicit representations~\cite{zhao2020physics} alike.

\subsection{Gaussian Mixtures in Rendering}

Gaussians or Gaussian Mixture Models (GMMs) have been used extensively 
in Computer Graphics to represent signals due to their simplicity and
convenient mathematical properties.
Zhou et al.~\shortcite{zhou2008real} used isotropic Gaussians to 
speed up volumetric path tracing by caching volumetric radiance 
fields. Jakob et al.~\shortcite{jakob2011progressive} fit a hierarchical 
mixture of anisotropic Gaussians to approximate million of photons 
with significantly fewer parameters. GMMs have also been utilized 
more recently for learning guiding distributions during path tracing~\cite{vorba2014line} to 
reduce noise. Such GMMs can be optimized robustly using 
expectation maximization even with noisy training samples. 
The convenient property of GMMs is that they can be used
for product importance sampling when the both the known and learned
components are available as Gaussians, resulting in a Gaussian product distribution~\cite{herholz2016product} that predicts reflected
light with high accuracy.

\subsection{Implicit Neural Rendering}

Neural networks were demonstrated to fit well to radiance fields of real~\cite{mildenhall2021nerf} and virtual scenes~\shortcite{ren2013global} when handed scene coordinates in the right format. \NEW{Networks have also been used to estimate the zero-level set of signed distance functions for neural surface reconstruction~\cite{wang2021neus}.}
After training, radiance or distance information is stored implicitly within the network's weights and biases. For every query point, an inference pass is required to retrieve the predicted value.
Full-image generation via neural networks~\cite{eslami2018neural} conditioned on observations from a small set of viewpoints
and decomposition of image-space scene information into smaller, less entangled representations~\cite{granskog2020compositional} for alternative renditions have also been studied for neural rendering.
Diolatzis et al.~\shortcite{diolatzis2022active} 
demonstrate a generator network for variable radiance fields based on 
explicit encoding of scene states. They achieve adaptive training data sampling by a guiding scheme that identifies elusive effects such as caustics or transparency. Small fully-fused neural networks that can be trained and evaluated in real time were popularized by Mueller et al.~\cite{muller2021real} for dynamic radiance caching and surface representations.
NeRF~\cite{mildenhall2021nerf} demonstrated that neural radiance field paired with volume ray casting can be fit to real-world view points and generate novel views with high quality. \NEW{In this application the encoding~\cite{tancik2020fourier} and parameterization of the inputs can have significant impact on quality. Ref-NeRF~\cite{verbin2022refnerf} replaced outgoing radiance with reflected radiance for improved reflections and view dependent effects.}

\subsection{Hybrid \& Explicit Representations}

The findings of neural radiance fields were subsequently
reproduced with more efficient hybrid and explicit
representations, where data structures tailored to the
spatio-angular domain are optimized to store some
or all of the information needed to reproduce training
views and synthesize novel views.
Octrees were applied to signed distance fields~\cite{takikawa2021neural} 
and radiance fields~\cite{yu2021plenoctrees}, omitting any neural components while retaining the iterative training scheme for volume ray casting samples. 
Dense grids~\cite{sun2022direct}, and more memory-efficient 
hash grids~\cite{muller2022instant} jointly optimized with small networks were shown effective and 
led to hybrid representations outperforming pure neural representations.
In generative adversarial networks for objects such as faces, 
triplane encodings~\cite{chan2022efficient} proved effective for many types of geometry despite their 2D nature. \NEW{As Radial Basis Functions~\cite{carr2001reconstruction, dinh2002reconstructing} have been shown in the past to be an efficient representation for 3D reconstruction, Neural Radial Basis functions were proposed in~\cite{chen2023neurbf}. They offered increased spatial adaptivity and outperformed grid-based approaches but extension to N-D would require a unified initialization and refinement scheme.} 
Our approach is inspired by the recent 3D Gaussian Splatting (3DGS) 
method~\cite{kerbl20233d} which demonstrates both impressive quality and 
performance for novel view synthesis of real world scenes using lower-dimensional GMMs, particularly due to their effective empty space skipping by way of direct projection to the image plane, paired with an efficient GPU rasterization implementation. \NEW{Such 3D Gaussian mixtures also can be turned into meshes as in~\cite{guedon2023sugar} for improved editability.}

\section{Method}

\begin{figure*}[!h]
	\includegraphics[width=\linewidth]{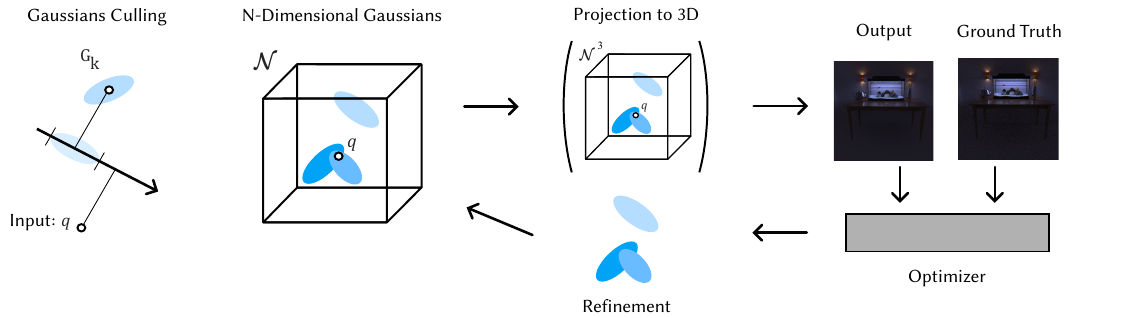}
	\caption{
		\label{fig:overview}
		Our optimization receives a number of query points q of N dimensionality as input. For these given points we estimate which Gaussians can be discarded safely through our N-Dimensional culling. With the remaining ones we evaluate for each q our Gaussian mixture either in N dimensions for surface radiance fields or by first projecting the Gaussians to 3D. Our optimization converges to high quality while it also controlling the introduction of new Gaussians via our Optimization-Controlled Refinement.
	}
\end{figure*}

In the following we describe the key aspects of our method,
which allow us to efficiently arrive at a compact explicit representation of density functions that are defined on high-dimensional input domains.
For our optimized target representation, we use Gaussian mixture models which fit well to highly anisotropic distributions of high-dimensional data.
Specifically, we optimize the shape and position of $n$ $N$-Dimensional Gaussian components jointly with learnable color values $c$ and an application-dependent parameter $a$ for brightness or opacity.

A key ingredient to efficient training is our culling step that omits any computations relating to Gaussians that are irrelevant to any current query point, by way of a Locality Sensitive Hashing-inspired approach that bounds their projections onto random unit vectors.
For adaptive refinement of our optimized representation, previously proposed schemes with explicit splitting and merging heuristics become difficult to tune to the sparse distributions of high-dimensional reference data. We introduce an optimizer-controlled refinement scheme, which smoothly fades in additional dependent Gaussians to enhance quality and add details where necessary.

\subsection{N-Dimensional Gaussian Parameterization}

We parameterize each of our $N$-Dimensional Gaussian components through its mean $\m: \mathbb{R}^N$ and its full covariance $\cov: \mathbb{R}^{N^2}$:
\begin{equation}
  G_{\cov}(\x-\m) = e^{-\frac{1}{2}(\x-\m)^T\cov^{-1}(\x-\m)} 
\end{equation}

\noindent The scale and rotation parameterization used in previous work~\cite{kerbl20233d} is problematic when we operate in $N$ dimensions, as rotations become increasingly difficult to describe. Instead, we opt for a Cholesky decomposition of $\cov$:
\begin{equation}
  \cov = \tril \tril^T
  \label{eq:decomp}
\end{equation}

\noindent where $\tril$ is a lower triangular matrix which we optimize directly.
We use an exponential activation function for the diagonal elements $\tril[i, i](x)=e^x$ to ensure positive-definiteness and stabilize training. For the lower triangular elements we restrict them to be in the -1 to 1 range using a sigmoid $\tril[i, j](x)=2 \cdot \text{sigmoid}(x)-1: i<j$.
Note that up to unitary transformations, the matrix~$\tril$ uniquely transforms a standard normal distribution~$G_{\ident}$ to a fitted Gaussian component~$G_{\cov=\tril \tril^T}$, which makes the decomposition well-suited for hierarchical transformations of Gaussians as used in our smooth optimization-controlled refinement.

\subsection{N-Dimensional Gaussians Culling}

In order to achieve acceptable performance during training and inference, we must avoid evaluating all the $N$-D Gaussians for each query point.
If all $N$-3 inputs except world position are constant for a single image, culling is straight-forward as global conditioning to the current values $\x_c = {x_4, \ldots, x_N}$ followed by standard spatial culling as in Kerbl et al.~\shortcite{kerbl20233d}.
%
In the general case that we aim at, however, inputs may vary between pixels (e.g. material-dependent fitting such as for surface roughness or refraction indices), requiring effective high-dimensional culling before costly projection and evaluation per pixel.

Our approach to culling is inspired by Locality Sensitive Hashing, which is sometimes used for high-dimensional approximate nearest neighbor search. The intuition behind LSH is that projection onto random vectors within the high-dimensional data space provide valuable information for estimating and bounding proximity.
We similarly project our Gaussian components onto random unit vectors to quickly discard components for which the query point falls outside their projected footprint.
%
Given a query point $\q: \mathbb{R}^N$ and an $N$-D Gaussian, we project both its mean $\m$ and covariance $\cov$ onto $k$ randomly sampled vectors $\rvec: \mathbb{R}^N$. The projections of the query point and the Gausssian mean are:
\begin{equation}
  q_{\rvec} = \q^T\rvec, \quad m_{\rvec} = \m^T\rvec
\end{equation}

\noindent To take the Gaussian's anisotropy into account we also project the covariance $\cov$ on the vector $\rvec$:
\begin{equation}
  \sigma^2_{\rvec} = \rvec^T\cov\rvec
\end{equation}

\noindent With this projection we can safely cull any Gaussian if:
\begin{equation}
  | q_{\rvec} - m_{\rvec} | < 3\sigma_{\rvec}
\end{equation}

\noindent i.e. if the query projection is outside the project $3\sigma$-confidence interval. Doing this for all pixels would be costly, instead we do it for each tile of 16 by 16 pixels, bounding the query point by their spatial extent.
In this way our culling is conservative, i.e. it will never discard a Gaussian that should have been evaluated.

\subsection{Optimization-Controlled Refinement}

\begin{figure}[b]
  \includegraphics[width=\linewidth]{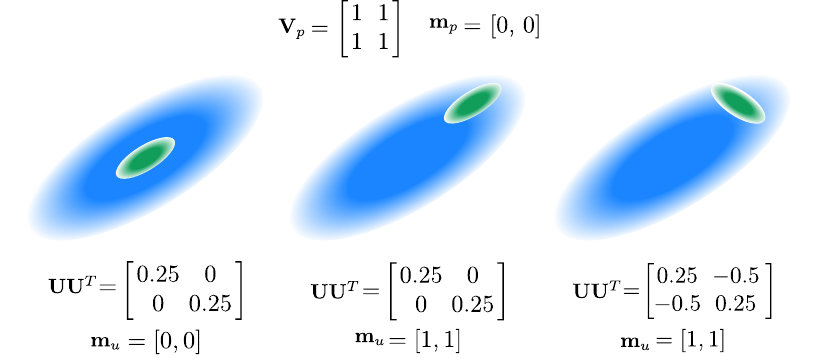}
  \caption
  {
    We visualize the relationship between parent (blue) and child (green) Gaussians for different covariance matrices and means.
  }
  \label{fig:sub_gs}
\end{figure}

In the following, we tackle the problem of refining our explicit Gaussian mixture representation in the face of unknown domain-specific splitting or merging heuristics (as previously proposed for e.g. spatio-angular domains~\cite{krivanek2006making,ruppert2020robust,kerbl20233d}). Note that heuristics based on auxiliary statistical estimates become increasingly unreliable with increasing number of dimensions due to increased scarcity of training samples w.r.t. any respective combinations of parameter values.
In order to arrive at a general, purely optimizer-controlled solution, we instead regularly introduce small amounts of additional capacity to our representation in the form of nested Gaussian components with a dependency structure that constrains their use: During training, by default each initial or newly materialized Gaussian component contains an additional child Gaussian with negligible randomized contribution, and spanning the full parent Gaussian. In order to avoid redundant fitting of parent mixture components, the child Gaussian~$G_{\covc}$ is defined in the reference frame of its parent Gaussian~$G_{\covp}$ by relative parameters~$\triu, \m_u$, thus receiving all parent updates by default:\CHANGED{Aligned naming conventions with the parent child naming scheme. }
\begin{align}
  \covc&=\tril\triu(\tril\triu)^T,\quad \covp=\tril\tril^T,
  \\
  \m_c &= \tril\m_u + \m_p.
  \label{eq:cov_dep}
\end{align}

\noindent We visualize this relationship in Figure~\ref{fig:sub_gs}. Thus, divergence of parent and child Gaussians is caused only in the need of \NEW{introducing new details which cannot be represented just by the parent Gaussian.} If such a case arises, \NEW{meaning the opacity/brightness of the child Gaussian is increased by the optimizer crossing a threshold of non-negligible contribution, the child Gaussian is materialized. This results in both parent and child becoming independent entities} and each is assigned a new default-initialized child Gaussian. Note that since we also parameterize child covariance in terms of a local lower triangular matrix~$\triu$, the materialized child Gaussian is again parameterized by a lower triangular matrix~$\tril\triu$ as the product of two such matrices preserves this property.

\paragraph{Practical Observations}
In this scheme, enforcing a dependency between the parent and child Gaussians is an important factor. Without this relationship the child Gaussians tend to move outside the parent Gaussian and become used redundantly. To avoid this, the covariance of the parent should impact the shape and mean of the child from the start to guide the optimization into using it only to enhance the parent Gaussian.

\begin{figure}[!h]
  \includegraphics[width=\linewidth]{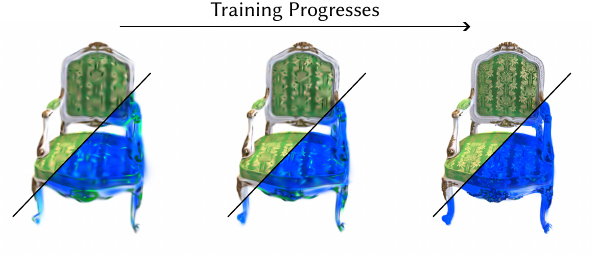}
  \caption
  {
    Our refinement scheme allows the optimizer to choose where to introduce new Gaussians. In the figure, at different parts of the training procedure, we show in blue the already existing Gaussians and in green the ones that the optimizer has chosen to utilize. Notice the dependent Gaussians adding details to the chair texture. 
  }
  \label{fig:refinement}
\end{figure}

Note that we avoid explicit subdivision in our refinement. In typical hierarchical subdivision schemes, finding the direction of the best split axis becomes increasingly complex. In our scheme, the iterative optimization is responsible for driving any added capacity of the respresentation into the right direction. In Figure~\ref{fig:refinement} we show an example of how the refinement introduces details using dependent Gaussians.

\paragraph{Practical Initialization and Refinement Scheduling}
We fit our representation in phases between refinements of about 300 iterations each, such that all Gaussians stabilize towards the end of each phase before any existing child Gaussians are materialized and new children added.
In the initial phase, we expect all Gaussians to move and change a lot, therefore we omit any nested child Gaussians until the initial state is sufficiently converged. Afterwards, the first set of child Gaussians are introduced and quickly start introducing some of the finer-grained effects and details. In order to determine which child Gaussians should be materialized after each phase, we check their brightness resp. opacity (depending on the application) to see if they are being utilized, in our experiments we use a threshold $t=0.1$ for opacity and $t=0.01$ for brightness. Different values of this threshold control how easily we want new Gaussians to be introduced versus forcing the optimizer to make do with fewer.
Note that when we introduce new child Gaussians, the overall values represented by the mixture do not change significantly, since we avoid any explicit splitting, subdivision, or other perturbation as used in previous work~\cite{kerbl20233d}. This avoids spiking in the loss and domain-specific tuning of heuristics. The contribution threshold $t$ is directly related to the expected output and thus an intuitive hyperparameter overall.

\section{Applications}

Our method offers a powerful representation for fitting high dimensional functions with strong anisotropy dependencies between dimensions. Such dependencies occur between position and direction for reflections or between position and time in time varying appearance. We apply our method to two different problems, global illumination of variable synthetic scenes and novel view synthesis with an emphasis on capturing view-dependent effects. We focus on these scenarios as they typically offer sufficiently dense capture data to allow for the optimization of high-dimensional functions. We leave generalization with sparser captures for future work.

\subsection{Global Illumination with Variability}

Neural networks have been utilized to shade synthetic scenes conditioned on fast to render G-Buffers~\cite{granskog2020compositional, diolatzis2022active}. When we apply our $N$-Dimensional Gaussians as a replacement for the generator networks used in these methods we achieve high quality within just a few minutes of training. 

In this scenario, geometry and material information are given through the G-Buffers, and each channel of these buffers provides a dimension for our Gaussians. We evaluate all Gaussians on the surfaces of the synthetic scene. More specifically the dimensionality of the Gaussians equals to the G-Buffers' dimensions plus any additional variable dimensions. The geometry and material information include 10 dimensions: world position $xyz$, view direction $xyz_{dir}$, albedo $rgb$ and roughness $r$. Any variable element of the scene at a given image has a specific state which is normalized using its variable range. For example, if an object can move from $x=-2$ to $x=2$, when it is at $x=0$ this variable dimension will have a value $v=0.5$. We concatenate these variable dimensions to the ones from the G-Buffers.

Since the geometry is given through G-Buffers the Gaussians have no opacity parameter, the parameter $\alpha$ controls instead the brightness of their color $c$. In other words for a given point on a surface the color computed from $k$ Gaussians is:
\begin{equation}
  c(\x) = G_{\cov_1}(\x-\m_1) \alpha_1 c_1 + \ldots + G_{\cov_k}(\x-\m_k) \alpha_k c_k 
\end{equation}

\noindent Since we train our method on high dynamic range renderings, we use an exponential activation for the brightness $\alpha$ and a sigmoid for the color $c$. Our loss function is a relative L2 loss~\cite{lehtinen2018noise2noise}, since the ground truth Monte Carlo renderings of our scenes have some residual noise. We use Mitsuba 3~\cite{Jakob2020DrJit} and its Python bindings to render our scene ground truths and G-Buffers. To reduce the time generating high quality datasets for each scene we train on 256 by 256 images and during rendering use higher resolution 800 by 800 G-Buffers. 

To achieve high performance during training and inference, we analytically compute the gradients for our trainable parameters and implement their calculations in Taichi~\cite{hu2019taichi} kernels.

The scenes we showcase have different amounts of variability. In the Bathroom scene we can change the camera and inspect the rough reflections of the geometry even though the Gaussians are evaluated on the surface of the transparent bathroom door. In the Living Room scene we can change the position of the light source while changing viewpoint. Finally, the Aquarium scene offers some complex multi bounce effects through the aquarium and the water, all which are simulated simply with anisotropy. In this scene we can also control the position of the fish. We showcase these results in Figure~\ref{fig:teaser}, Figure~\ref{fig:comp_mitsuba} and in our video.

\subsection{Volumetric Radiance Fields}

Another application where neural networks have been used extensively is novel view synthesis, with methods like NeRF training a network to store the volumetric radiance field of a scene. One common bias in both implicit and explicit methods for novel view synthesis is the reduced representation power in the view direction component compared to the spatial one. Except for specialized methods, this leads to high-frequency reflections being reconstructed poorly or as duplicate geometry. Such approaches fail in the case of complex anisotropic reflections and leading to low-quality reconstructions.

In contrast, a 6D spatio-angular representation constructed using our method can have as much adaptivity in the angular as in the spatial domain, and anisotropic relationships between the two domains are modeled explicitly. We demonstrate our method on such challenging scenes as the ones in the dataset provided by~\cite{Wizadwongsa2021NeX}. We show this application to demonstrate the versatility and robustness of our method. To apply our method to this scenario, we only modify the rendering step, i.e. we apply a projection from N-D to 3D to our Gaussians \NEW{(details of the projection are included in supplemental)} and then use the splatting process from 3DGS. Despite the quite different application of rendering real world scenes our N-D parameterization and dependent refinement scheme works well, adapting to details and giving high quality reconstruction in both diffuse and view dependent effects. We showcase the results in Figure~\ref{fig:teaser}, Figure~\ref{fig:comp_3dgs} and in the video.

\section{Comparisons \& Ablations}

To evaluate the effectiveness and robustness of our method we compare with implicit, hybrid and explicit methods in our two different applications. 

Specifically for the shading of variable scenes we train a Pixel Generator network (a MLP network with skip connections) that has been used in previous methods~\cite{granskog2020compositional,diolatzis2022active}, and we use TinyCudaNN~\cite{tiny-cuda-nn} to train a hash grid and a small MLP decoder. 

For novel view synthesis on scenes with hard to render anisotropic reflections we compare against 3D Gaussian Splatting for comparable training time. 

\paragraph{Comparison for Synthetic Scenes}

We train a deep neural network (8 layers of 512 features) with the Pixel Generator architecture used and an implementation of the hash grid encoding for comparison.

\begin{figure*}[!h]
	\includegraphics[width=\linewidth]{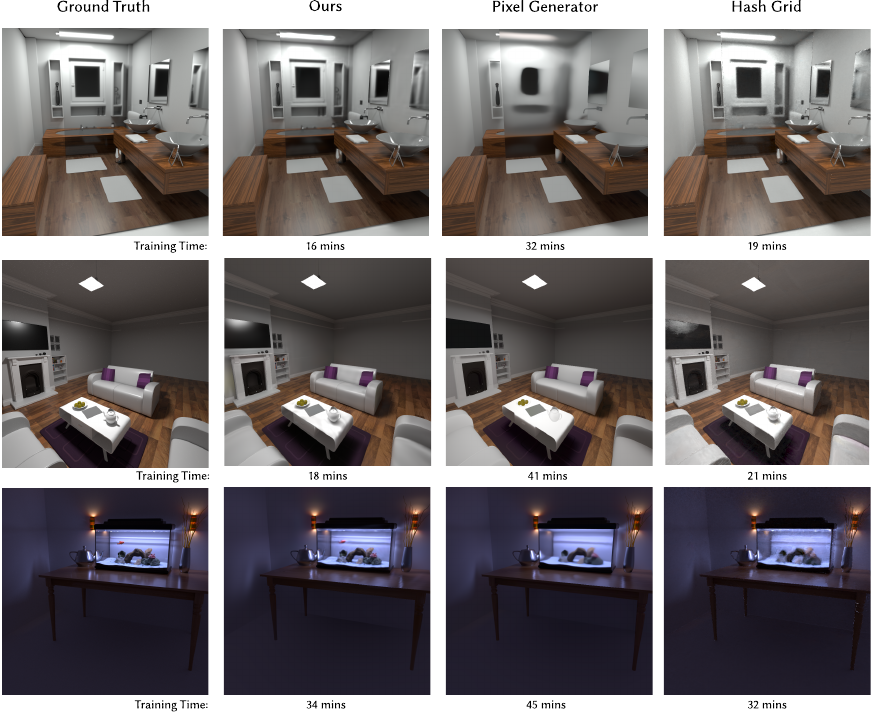}
	\caption{
		\label{fig:comp_mitsuba}
		\CHANGED{Moved figure to comply with the guidelines. Added noise free ground truths.}We compare with an implicit (Pixel Generator) and hybrid (Hash Grid) method for our application of shading variable synthetic scenes. Our method achieves high quality in a few minutes of training while enabling fast rendering and interactivity.
	}
\end{figure*}

For the hash encoding we parameterize xyz using the hash grid, spherical harmonics for the view direction and an identity encoding for the rest of the dimensions, all of which is decoded by a 64 feature 2 layers MLP. Despite our method not using any explicit domain knowledge of the inputs to use different encodings, it achieves superior quality as showcased in Figure~\ref{fig:comp_mitsuba}. In contrast, the shallow MLP decoder is unable to handle the collisions of the hash grid while taking into account the extra 7 input dimensions. 

The Pixel Generator is quite a big network and one that needs close to tens of hours of training to converge. For similar training times it learns a blurry representation compared to our explicit optimization (Figure~\ref{fig:comp_mitsuba}).

We give our quantitative evaluations in Table~\ref{table:comp_mitsuba} for all 3 methods and scenes. \NEW{In Table~\ref{table:comp_lsh} we also report the iteration timings for each method. Note that the Hash Grid baseline utilizes Python bindings to optimized C++ code compared to our method which uses purely Python.}

\begin{table}[!h]
  \caption
  {
    In this table we report Mean Absolute Percentage Error (MAPE) for each method in our synthetic scenes dataset.
  }
  \includegraphics[width=\linewidth]{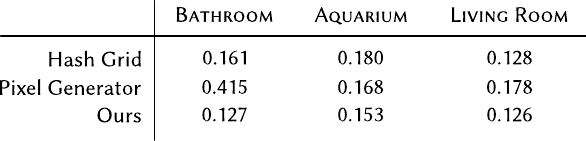}
  \label{table:comp_mitsuba}
\end{table}

\paragraph{Comparison for Specular Real World Scenes}

\NEW{In this application we initialize the spatial component and color of our Gaussians, similarly to 3DGS, using Structure-from-Motion (SfM)~\cite{snavely2006photo} when available. Since these points can create erroneous Gaussians, after initialization we discard any that have very low opacity ($a$ < 0.1). Compared to 3DGS, our method doesn't require regular discarding or resetting of opacity.}

\NEW{We compare against 3DGS as it is conceptually close to our method, Instant NGP as a state of the art hybrid method, and to NeX as an implicit method with a focus on specular effects. We train 3DGS and INGP for similar training times using the respective publicly available code. As NeX requires tens of hours to train we use the numbers reported in their paper and train our method in the same (lower) resolution.}

\begin{table}[!h]
  \caption
  {
    \CHANGED{ Added INGP and NeX metrics and updated metrics for our method and 3DGS after solving an issue with off-center cameras which increased PSNR in both ours and 3DGS.}\NEW{Quantitative evaluation of our method against 3DGS, Instant NGP and NeX for our two scenes. We show training resolution, training times and PSNR for all methods. Nex metrics are provided from their paper.}
  }
  \includegraphics[width=\linewidth]{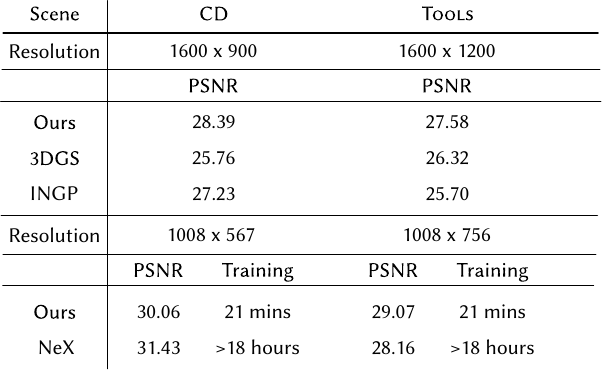}
  \label{table:comp_shiny}
\end{table}  

\begin{figure*}[!h]
	\includegraphics[width=\linewidth]{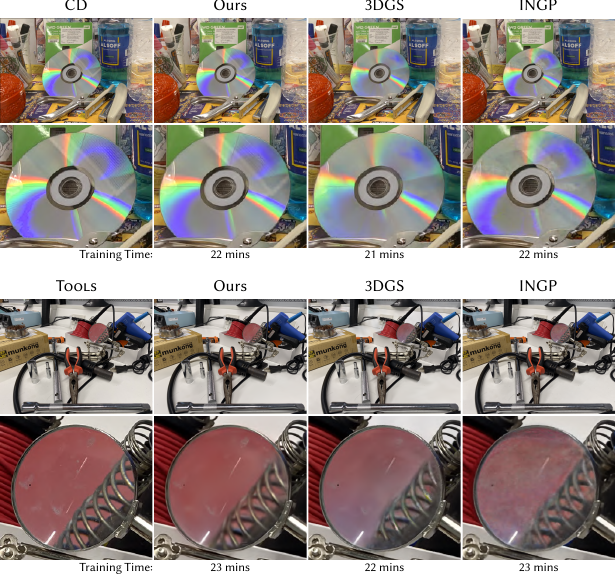}
	\caption{
		\label{fig:comp_3dgs}
		\CHANGED{Moved figure after references to comply with the guidelines. Added INGP comparison and updated our results and 3DGS corresponding to the new metrics in Table~\ref{table:comp_shiny}. }\NEW{Qualitative results of our method compared 3DGS and Instant-NGP for two different scenes with complex view dependent effects.}
	}
\end{figure*}

As we show in Figure~\ref{fig:comp_3dgs} for scenes with highly anisotropic reflections which cannot be faked by floating geometry the spherical harmonics encoding used in 3DGS is not enough. \NEW{Instant NGP fares better in reproducing some specular effects, but the capacity of the small network cannot handle all the collisions efficiently, leading to blurring and fogginess.} Our method learns correctly to optimize the anisotropy between world position and viewing direction which results in reflections actually moving in world space as we move the camera. In our video please appreciate that many of the reflections are parameterized by a few Gaussians (demonstrated through the color-coded Gaussians) resulting in fewer ones necessary to achieve higher quality. \NEW{These observations are also reflected in PSNR in Table~\ref{table:comp_shiny}. Since our method utilizes fewer Gaussians, it also has overall shorter evaluation times. In Table~\ref{table:comp_eval} we show the inference times for our method and 3DGS using the Python implementation of the splatting renderer provided by the 3DGS authors. Note that 3DGS also provides a C++ renderer, part of~\cite{sibr2020}, which provides much better performance than the Python version but our 6D parameterization could be integrated within the same framework to profit from similar performance gains.}

\begin{table}[!h]
  \caption
  {
    \CHANGED{ Added evaluation (inference) time compared to 3DGS for the two 6D scenes \textsc{CD} and \textsc{Tools}} \NEW{We report the number of Gaussians at the end of training for our method and 3DGS as well as the inference time for both methods using the Python splatting renderer of~\cite{kerbl20233d}.}
  }
  \includegraphics[width=\linewidth]{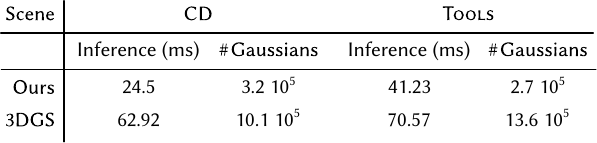}
  \label{table:comp_eval}
\end{table}  

\NEW{Compared to NeX we demonstrate similar PSNR, on average, within half an hour of training compared to the tens of hours required by NeX. Also note that training NeX in the same higher resolution as the previous results required more memory than the 24 GB of available GPU memory in our tests.}

\paragraph{LSH Ablation}

Our tile-based high dimensional culling using projections allows us to train and render fast our mixtures. We showcase these advantages against brute force evaluation in Table~\ref{table:comp_lsh} for the synthetic scenes application. During training we use lower resolution images and in that case the 16 by 16 tiles cover a larger area resulting in fewer Gaussians culled. Still avoiding the back propagation step for these Gaussians saves a considerable amount of time for each training iteration. 

At inference time each tile covers a small area and as a result a big percentage of the Gaussians can be discarded resulting in three times faster evaluation.

\begin{table}[!h]
  \caption
  {
    \NEW{Iteration time (during training and inference) of our method (with and without LSH culling), the Pixel Generator and the Hash Grid baselines for the \textsc{Bathroom} scene}.
  }
  \includegraphics[width=\linewidth]{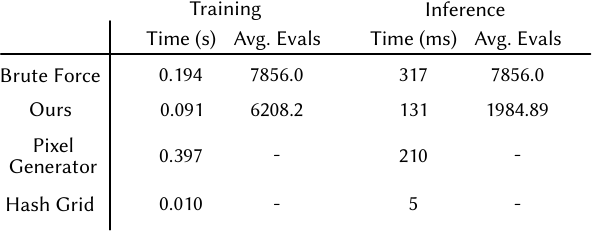}
  \label{table:comp_lsh}
\end{table}

\begin{figure}[!h]
  \includegraphics[width=\linewidth]{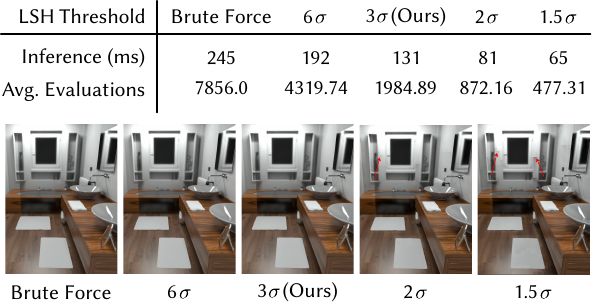}
  \caption
  {
    \CHANGED{Added new figure which shows the impact of different LSH thresholds on image quality.}\NEW{Demonstration of the impact to image quality and inference time for different values of LSH threshold. With red arrows we point out the artifacts that appear when the threshold is set to lower values than our choice of $3\sigma$.}
  }
  \label{fig:comp_lsh_sigma}
\end{figure}

\NEW{To test the impact of the chosen confidence interval on performance and image quality we experiment with different values including our choice of $3\sigma$ and report the results in Figure~\ref{fig:comp_lsh_sigma}.}
\NEW{Lower confidence values lead to more aggressive culling with lower average evaluations (more Gaussians discarded) and faster inference times. But in Figure~\ref{fig:comp_lsh_sigma} we observe that for values lower than the chosen $3\sigma$ threshold, blocky artifacts start to appear (highlighted by red arrows) due to the aggressive culling. This means that for these values the culling discards Gaussians which normally contributed to the image leading to artifacts.}

\section{Limitations and Future Work}

Naturally, any representation fitting to reference data is prone to overfitting and aliasing problems, to which our method is no exception. It is well-known that this challenge increases with the number of variables affecting the results~\cite[Sect. 5.1.3]{figueiredo2002unsupervised}, calling for according increases in training sample size. In our experiments fitting to real-world view points, we avoid overfitting our directionally adaptive representation to sparse view point sets using typical directional regularization~\cite{niemeyer2022regnerf}: Smoothness of directionally varying components is enforced in-between captured view points by randomly perturbing training directions to cover the full space. \NEW{Even with this regularization real-world 360 datasets are more challenging in their data distribution and capture quality, and it could require specific tuning/heuristics which we leave for future work.}

\NEW{Adapting our refinement scheme in the future to other covariance parameterizations, such as the one used in 3DGS could help reduce the amount of hyperparameters required. However such an adaptation is non-trivial due to additional conversions required in the materialization (composite covariance to quaternions + scaling) and would require extra care to not degrade performance overall.}

The computational scaling with respect to number of input dimensions differs for implicit neural and explicit Gaussian representations. In the plain Gaussian mixture parameterization used throughout this paper, storage cost per Gaussian component grows quadratically with the number of dimensions. More sparse parameterizations would be possible (e.g. trading full inter-dimensional adaptivity for fewer principal component axes), but are subject to future work. We note that while in contrast, neural representations may implicitly expose sub-quadratic scaling behavior, no such representations suitable for interactive rendering are currently easily extensible to high-dimensional domains.

The choice of representations and their parameterization has observable impact on results. Like in previous work~\cite{kerbl20233d}, the representation of discontinuities by Gaussian mixture results in somewhat scale-dependent sharpness, whereas other representations such as implicit neural models using ReLU activations are naturally well-suited to place truly discontinuous boundaries. Studying alternative distribution primitives as well as alternative distribution parameters for the best control, convergence, and compactness of resulting representations, could be interesting future work.
Capturing coherent motion of primarily visible objects is another interesting open issue for follow-up work. While in principle, our Gaussian components can be moved easily in the spatial domain, we note that such a manual animation would be in contrast to captured time-dependent signals: Even time-dependent evaluation of our representation is effectively a slice through a static high-dimensional Gaussian mixture, therefore our method does not yet implicitly solve correspondence-finding of coherently transformed objects. We expect interesting challenges for future high-dimensional explicit representations to reach this secondary objective.

\section{Conclusion}

We demonstrated feasibility of efficiently constructing compact explicit representations for the reconstruction of high-dimensional data in interactive computer graphics applications. In particular, we achieve similar or higher quality in our results when compared to previous methods, without leveraging domain knowledge that is restricted to lower-dimensional domains. We hope that our insights can inspire both explicit and neural implicit representations going forward to expand their scope to adaptivity in higher-dimensional input domains, either by way of improved fitting of explicit primitives or by way of derived parametric encodings.

The locality of parameter impact in explicit representations is a strong property not only for culling and efficiency purposes, but also in the sense of distributed storage capacity, providing good opportunities for an implicit adaptive scaling to larger input domains. In contrast, implicit neural representations commonly assign a certain fixed capacity to certain parts of the domain. Ultimately, however, both approaches will still require efficient large-scale culling mechanisms (e.g. using hierarchical schemes) accounting for high variability with respect to high-dimensional inputs, which is another interesting area of future work.

\begin{acks}
  We would like to thank the anonymous referees for their valuable comments and helpful suggestions. We also thank Laurent Belcour and Sebastian Herholz for their valuable input and suggestions. G. Kopanas was supported by ERC Advanced grant FUNGRAPH No. 788065~(\url{http://fungraph.inria.fr}).
\end{acks}

\bibliographystyle{ACM-Reference-Format}
\bibliography{ingi-paper}


\begin{thebibliography}{39}


\ifx \showCODEN    \undefined \def \showCODEN     #1{\unskip}     \fi
\ifx \showDOI      \undefined \def \showDOI       #1{#1}\fi
\ifx \showISBNx    \undefined \def \showISBNx     #1{\unskip}     \fi
\ifx \showISBNxiii \undefined \def \showISBNxiii  #1{\unskip}     \fi
\ifx \showISSN     \undefined \def \showISSN      #1{\unskip}     \fi
\ifx \showLCCN     \undefined \def \showLCCN      #1{\unskip}     \fi
\ifx \shownote     \undefined \def \shownote      #1{#1}          \fi
\ifx \showarticletitle \undefined \def \showarticletitle #1{#1}   \fi
\ifx \showURL      \undefined \def \showURL       {\relax}        \fi
\providecommand\bibfield[2]{#2}
\providecommand\bibinfo[2]{#2}
\providecommand\natexlab[1]{#1}
\providecommand\showeprint[2][]{arXiv:#2}

\bibitem[Andoni et~al\mbox{.}(2015)]%
        {andoni2015practical}
\bibfield{author}{\bibinfo{person}{Alexandr Andoni}, \bibinfo{person}{Piotr
  Indyk}, \bibinfo{person}{Thijs Laarhoven}, \bibinfo{person}{Ilya
  Razenshteyn}, {and} \bibinfo{person}{Ludwig Schmidt}.}
  \bibinfo{year}{2015}\natexlab{}.
\newblock \showarticletitle{Practical and optimal LSH for angular distance}.
\newblock \bibinfo{journal}{\emph{Advances in neural information processing
  systems}}  \bibinfo{volume}{28} (\bibinfo{year}{2015}).
\newblock


\bibitem[Bonopera et~al\mbox{.}(2020)]%
        {sibr2020}
\bibfield{author}{\bibinfo{person}{Sebastien Bonopera}, \bibinfo{person}{Peter
  Hedman}, \bibinfo{person}{Jerome Esnault}, \bibinfo{person}{Siddhant
  Prakash}, \bibinfo{person}{Simon Rodriguez}, \bibinfo{person}{Theo Thonat},
  \bibinfo{person}{Mehdi Benadel}, \bibinfo{person}{Gaurav Chaurasia},
  \bibinfo{person}{Julien Philip}, {and} \bibinfo{person}{George Drettakis}.}
  \bibinfo{year}{2020}\natexlab{}.
\newblock \bibinfo{title}{sibr: A System for Image Based Rendering}.
\newblock
\newblock
\urldef\tempurl%
\url{https://sibr.gitlabpages.inria.fr/}
\showURL{%
\tempurl}


\bibitem[Carr et~al\mbox{.}(2001)]%
        {carr2001reconstruction}
\bibfield{author}{\bibinfo{person}{Jonathan~C Carr}, \bibinfo{person}{Richard~K
  Beatson}, \bibinfo{person}{Jon~B Cherrie}, \bibinfo{person}{Tim~J Mitchell},
  \bibinfo{person}{W~Richard Fright}, \bibinfo{person}{Bruce~C McCallum}, {and}
  \bibinfo{person}{Tim~R Evans}.} \bibinfo{year}{2001}\natexlab{}.
\newblock \showarticletitle{Reconstruction and representation of 3D objects
  with radial basis functions}. In \bibinfo{booktitle}{\emph{Proceedings of the
  28th annual conference on Computer graphics and interactive techniques}}.
  \bibinfo{pages}{67--76}.
\newblock


\bibitem[Chan et~al\mbox{.}(2022)]%
        {chan2022efficient}
\bibfield{author}{\bibinfo{person}{Eric~R Chan}, \bibinfo{person}{Connor~Z
  Lin}, \bibinfo{person}{Matthew~A Chan}, \bibinfo{person}{Koki Nagano},
  \bibinfo{person}{Boxiao Pan}, \bibinfo{person}{Shalini De~Mello},
  \bibinfo{person}{Orazio Gallo}, \bibinfo{person}{Leonidas~J Guibas},
  \bibinfo{person}{Jonathan Tremblay}, \bibinfo{person}{Sameh Khamis},
  {et~al\mbox{.}}} \bibinfo{year}{2022}\natexlab{}.
\newblock \showarticletitle{Efficient geometry-aware 3D generative adversarial
  networks}. In \bibinfo{booktitle}{\emph{Proc. IEEE CVPR}}.
  \bibinfo{pages}{16123--16133}.
\newblock


\bibitem[Chen et~al\mbox{.}(2023)]%
        {chen2023neurbf}
\bibfield{author}{\bibinfo{person}{Zhang Chen}, \bibinfo{person}{Zhong Li},
  \bibinfo{person}{Liangchen Song}, \bibinfo{person}{Lele Chen},
  \bibinfo{person}{Jingyi Yu}, \bibinfo{person}{Junsong Yuan}, {and}
  \bibinfo{person}{Yi Xu}.} \bibinfo{year}{2023}\natexlab{}.
\newblock \showarticletitle{NeuRBF: A Neural Fields Representation with
  Adaptive Radial Basis Functions}. In \bibinfo{booktitle}{\emph{Proceedings of
  the IEEE/CVF International Conference on Computer Vision (ICCV)}}.
  \bibinfo{pages}{4182--4194}.
\newblock


\bibitem[Dinh et~al\mbox{.}(2002)]%
        {dinh2002reconstructing}
\bibfield{author}{\bibinfo{person}{Huong~Quynh Dinh}, \bibinfo{person}{Greg
  Turk}, {and} \bibinfo{person}{Greg Slabaugh}.}
  \bibinfo{year}{2002}\natexlab{}.
\newblock \showarticletitle{Reconstructing surfaces by volumetric
  regularization using radial basis functions}.
\newblock \bibinfo{journal}{\emph{IEEE transactions on pattern analysis and
  machine intelligence}} \bibinfo{volume}{24}, \bibinfo{number}{10}
  (\bibinfo{year}{2002}), \bibinfo{pages}{1358--1371}.
\newblock


\bibitem[Diolatzis et~al\mbox{.}(2022)]%
        {diolatzis2022active}
\bibfield{author}{\bibinfo{person}{Stavros Diolatzis}, \bibinfo{person}{Julien
  Philip}, {and} \bibinfo{person}{George Drettakis}.}
  \bibinfo{year}{2022}\natexlab{}.
\newblock \showarticletitle{Active exploration for neural global illumination
  of variable scenes}.
\newblock \bibinfo{journal}{\emph{ACM Trans. Graph.}} \bibinfo{volume}{41},
  \bibinfo{number}{5} (\bibinfo{year}{2022}), \bibinfo{pages}{1--18}.
\newblock


\bibitem[Eslami et~al\mbox{.}(2018)]%
        {eslami2018neural}
\bibfield{author}{\bibinfo{person}{SM~Ali Eslami}, \bibinfo{person}{Danilo
  Jimenez~Rezende}, \bibinfo{person}{Frederic Besse}, \bibinfo{person}{Fabio
  Viola}, \bibinfo{person}{Ari~S Morcos}, \bibinfo{person}{Marta Garnelo},
  \bibinfo{person}{Avraham Ruderman}, \bibinfo{person}{Andrei~A Rusu},
  \bibinfo{person}{Ivo Danihelka}, \bibinfo{person}{Karol Gregor},
  {et~al\mbox{.}}} \bibinfo{year}{2018}\natexlab{}.
\newblock \showarticletitle{Neural scene representation and rendering}.
\newblock \bibinfo{journal}{\emph{Science}} \bibinfo{volume}{360},
  \bibinfo{number}{6394} (\bibinfo{year}{2018}), \bibinfo{pages}{1204--1210}.
\newblock


\bibitem[Figueiredo and Jain(2002)]%
        {figueiredo2002unsupervised}
\bibfield{author}{\bibinfo{person}{M.A.T. Figueiredo} {and}
  \bibinfo{person}{A.K. Jain}.} \bibinfo{year}{2002}\natexlab{}.
\newblock \showarticletitle{Unsupervised learning of finite mixture models}.
\newblock \bibinfo{journal}{\emph{IEEE Transactions on Pattern Analysis and
  Machine Intelligence}} \bibinfo{volume}{24}, \bibinfo{number}{3}
  (\bibinfo{year}{2002}), \bibinfo{pages}{381--396}.
\newblock
\urldef\tempurl%
\url{https://doi.org/10.1109/34.990138}
\showDOI{\tempurl}


\bibitem[Granskog et~al\mbox{.}(2020)]%
        {granskog2020compositional}
\bibfield{author}{\bibinfo{person}{Jonathan Granskog}, \bibinfo{person}{Fabrice
  Rousselle}, \bibinfo{person}{Marios Papas}, {and} \bibinfo{person}{Jan
  Nov{\'a}k}.} \bibinfo{year}{2020}\natexlab{}.
\newblock \showarticletitle{Compositional neural scene representations for
  shading inference}.
\newblock \bibinfo{journal}{\emph{ACM Trans. Graph.}} \bibinfo{volume}{39},
  \bibinfo{number}{4} (\bibinfo{year}{2020}), \bibinfo{pages}{135--1}.
\newblock


\bibitem[Gu{\'e}don and Lepetit(2023)]%
        {guedon2023sugar}
\bibfield{author}{\bibinfo{person}{Antoine Gu{\'e}don} {and}
  \bibinfo{person}{Vincent Lepetit}.} \bibinfo{year}{2023}\natexlab{}.
\newblock \showarticletitle{Sugar: Surface-aligned gaussian splatting for
  efficient 3d mesh reconstruction and high-quality mesh rendering}.
\newblock \bibinfo{journal}{\emph{arXiv preprint arXiv:2311.12775}}
  (\bibinfo{year}{2023}).
\newblock


\bibitem[Herholz et~al\mbox{.}(2016)]%
        {herholz2016product}
\bibfield{author}{\bibinfo{person}{Sebastian Herholz}, \bibinfo{person}{Oskar
  Elek}, \bibinfo{person}{Ji{\v{r}}{\'\i} Vorba}, \bibinfo{person}{Hendrik
  Lensch}, {and} \bibinfo{person}{Jaroslav K{\v{r}}iv{\'a}nek}.}
  \bibinfo{year}{2016}\natexlab{}.
\newblock \showarticletitle{Product importance sampling for light transport
  path guiding}. In \bibinfo{booktitle}{\emph{Comp. Graph. Forum}},
  Vol.~\bibinfo{volume}{35}. Wiley Online Library, \bibinfo{pages}{67--77}.
\newblock


\bibitem[Hu et~al\mbox{.}(2019)]%
        {hu2019taichi}
\bibfield{author}{\bibinfo{person}{Yuanming Hu}, \bibinfo{person}{Tzu-Mao Li},
  \bibinfo{person}{Luke Anderson}, \bibinfo{person}{Jonathan Ragan-Kelley},
  {and} \bibinfo{person}{Fr{\'e}do Durand}.} \bibinfo{year}{2019}\natexlab{}.
\newblock \showarticletitle{Taichi: a language for high-performance computation
  on spatially sparse data structures}.
\newblock \bibinfo{journal}{\emph{ACM Trans. Graph.}} \bibinfo{volume}{38},
  \bibinfo{number}{6} (\bibinfo{year}{2019}), \bibinfo{pages}{1--16}.
\newblock


\bibitem[Jakob et~al\mbox{.}(2011)]%
        {jakob2011progressive}
\bibfield{author}{\bibinfo{person}{Wenzel Jakob}, \bibinfo{person}{Christian
  Regg}, {and} \bibinfo{person}{Wojciech Jarosz}.}
  \bibinfo{year}{2011}\natexlab{}.
\newblock \showarticletitle{Progressive expectation-maximization for
  hierarchical volumetric photon mapping}. In \bibinfo{booktitle}{\emph{Comp.
  Graph. Forum}}, Vol.~\bibinfo{volume}{30}. \bibinfo{pages}{1287--1297}.
\newblock


\bibitem[Jakob et~al\mbox{.}(2022)]%
        {Jakob2020DrJit}
\bibfield{author}{\bibinfo{person}{Wenzel Jakob}, \bibinfo{person}{Sébastien
  Speierer}, \bibinfo{person}{Nicolas Roussel}, {and} \bibinfo{person}{Delio
  Vicini}.} \bibinfo{year}{2022}\natexlab{}.
\newblock \showarticletitle{Dr.Jit: A Just-In-Time Compiler for Differentiable
  Rendering}.
\newblock \bibinfo{journal}{\emph{Transactions on Graphics (Proceedings of
  SIGGRAPH)}} \bibinfo{volume}{41}, \bibinfo{number}{4} (\bibinfo{date}{July}
  \bibinfo{year}{2022}).
\newblock
\urldef\tempurl%
\url{https://doi.org/10.1145/3528223.3530099}
\showDOI{\tempurl}


\bibitem[Kerbl et~al\mbox{.}(2023)]%
        {kerbl20233d}
\bibfield{author}{\bibinfo{person}{Bernhard Kerbl}, \bibinfo{person}{Georgios
  Kopanas}, \bibinfo{person}{Thomas Leimk{\"u}hler}, {and}
  \bibinfo{person}{George Drettakis}.} \bibinfo{year}{2023}\natexlab{}.
\newblock \showarticletitle{3D Gaussian Splatting for Real-Time Radiance Field
  Rendering}.
\newblock \bibinfo{journal}{\emph{ACM Trans. Graph.}} \bibinfo{volume}{42},
  \bibinfo{number}{4} (\bibinfo{year}{2023}).
\newblock


\bibitem[Kitaev et~al\mbox{.}(2020)]%
        {kitaev2020reformer}
\bibfield{author}{\bibinfo{person}{Nikita Kitaev}, \bibinfo{person}{{\L}ukasz
  Kaiser}, {and} \bibinfo{person}{Anselm Levskaya}.}
  \bibinfo{year}{2020}\natexlab{}.
\newblock \showarticletitle{Reformer: The efficient transformer}.
\newblock \bibinfo{journal}{\emph{arXiv preprint arXiv:2001.04451}}
  (\bibinfo{year}{2020}).
\newblock


\bibitem[Kriv{\'a}nek et~al\mbox{.}(2006)]%
        {krivanek2006making}
\bibfield{author}{\bibinfo{person}{Jaroslav Kriv{\'a}nek},
  \bibinfo{person}{Kadi Bouatouch}, \bibinfo{person}{Sumanta~N Pattanaik},
  {and} \bibinfo{person}{Jiri Zara}.} \bibinfo{year}{2006}\natexlab{}.
\newblock \showarticletitle{Making Radiance and Irradiance Caching Practical:
  Adaptive Caching and Neighbor Clamping.}
\newblock \bibinfo{journal}{\emph{Rendering Techniques}}
  \bibinfo{volume}{2006} (\bibinfo{year}{2006}), \bibinfo{pages}{127--138}.
\newblock


\bibitem[Lehtinen et~al\mbox{.}(2018)]%
        {lehtinen2018noise2noise}
\bibfield{author}{\bibinfo{person}{Jaakko Lehtinen}, \bibinfo{person}{Jacob
  Munkberg}, \bibinfo{person}{Jon Hasselgren}, \bibinfo{person}{Samuli Laine},
  \bibinfo{person}{Tero Karras}, \bibinfo{person}{Miika Aittala}, {and}
  \bibinfo{person}{Timo Aila}.} \bibinfo{year}{2018}\natexlab{}.
\newblock \showarticletitle{Noise2Noise: Learning image restoration without
  clean data}.
\newblock \bibinfo{journal}{\emph{arXiv preprint arXiv:1803.04189}}
  (\bibinfo{year}{2018}).
\newblock


\bibitem[Mildenhall et~al\mbox{.}(2021)]%
        {mildenhall2021nerf}
\bibfield{author}{\bibinfo{person}{Ben Mildenhall}, \bibinfo{person}{Pratul~P
  Srinivasan}, \bibinfo{person}{Matthew Tancik}, \bibinfo{person}{Jonathan~T
  Barron}, \bibinfo{person}{Ravi Ramamoorthi}, {and} \bibinfo{person}{Ren Ng}.}
  \bibinfo{year}{2021}\natexlab{}.
\newblock \showarticletitle{Nerf: Representing scenes as neural radiance fields
  for view synthesis}.
\newblock \bibinfo{journal}{\emph{Comm. ACM}} \bibinfo{volume}{65},
  \bibinfo{number}{1} (\bibinfo{year}{2021}), \bibinfo{pages}{99--106}.
\newblock


\bibitem[M\"uller(2021)]%
        {tiny-cuda-nn}
\bibfield{author}{\bibinfo{person}{Thomas M\"uller}.}
  \bibinfo{year}{2021}\natexlab{}.
\newblock \bibinfo{booktitle}{\emph{{tiny-cuda-nn}}}.
\newblock
\urldef\tempurl%
\url{https://github.com/NVlabs/tiny-cuda-nn}
\showURL{%
\tempurl}


\bibitem[M{\"u}ller et~al\mbox{.}(2022)]%
        {muller2022instant}
\bibfield{author}{\bibinfo{person}{Thomas M{\"u}ller}, \bibinfo{person}{Alex
  Evans}, \bibinfo{person}{Christoph Schied}, {and} \bibinfo{person}{Alexander
  Keller}.} \bibinfo{year}{2022}\natexlab{}.
\newblock \showarticletitle{Instant neural graphics primitives with a
  multiresolution hash encoding}.
\newblock \bibinfo{journal}{\emph{ACM Trans. Graph.}} \bibinfo{volume}{41},
  \bibinfo{number}{4} (\bibinfo{year}{2022}), \bibinfo{pages}{1--15}.
\newblock


\bibitem[M{\"u}ller et~al\mbox{.}(2021)]%
        {muller2021real}
\bibfield{author}{\bibinfo{person}{Thomas M{\"u}ller}, \bibinfo{person}{Fabrice
  Rousselle}, \bibinfo{person}{Jan Nov{\'a}k}, {and} \bibinfo{person}{Alexander
  Keller}.} \bibinfo{year}{2021}\natexlab{}.
\newblock \showarticletitle{Real-time neural radiance caching for path
  tracing}.
\newblock \bibinfo{journal}{\emph{ACM Trans. Graph.}} \bibinfo{volume}{40},
  \bibinfo{number}{4} (\bibinfo{year}{2021}), \bibinfo{pages}{1--16}.
\newblock


\bibitem[Niemeyer et~al\mbox{.}(2022)]%
        {niemeyer2022regnerf}
\bibfield{author}{\bibinfo{person}{Michael Niemeyer},
  \bibinfo{person}{Jonathan~T Barron}, \bibinfo{person}{Ben Mildenhall},
  \bibinfo{person}{Mehdi~SM Sajjadi}, \bibinfo{person}{Andreas Geiger}, {and}
  \bibinfo{person}{Noha Radwan}.} \bibinfo{year}{2022}\natexlab{}.
\newblock \showarticletitle{Regnerf: Regularizing neural radiance fields for
  view synthesis from sparse inputs}. In \bibinfo{booktitle}{\emph{Proceedings
  of the IEEE/CVF Conference on Computer Vision and Pattern Recognition}}.
  \bibinfo{pages}{5480--5490}.
\newblock


\bibitem[Ren et~al\mbox{.}(2013)]%
        {ren2013global}
\bibfield{author}{\bibinfo{person}{Peiran Ren}, \bibinfo{person}{Jiaping Wang},
  \bibinfo{person}{Minmin Gong}, \bibinfo{person}{Stephen Lin},
  \bibinfo{person}{Xin Tong}, {and} \bibinfo{person}{Baining Guo}.}
  \bibinfo{year}{2013}\natexlab{}.
\newblock \showarticletitle{Global illumination with radiance regression
  functions.}
\newblock \bibinfo{journal}{\emph{ACM Trans. Graph.}} \bibinfo{volume}{32},
  \bibinfo{number}{4} (\bibinfo{year}{2013}), \bibinfo{pages}{130--1}.
\newblock


\bibitem[Ruppert et~al\mbox{.}(2020)]%
        {ruppert2020robust}
\bibfield{author}{\bibinfo{person}{Lukas Ruppert}, \bibinfo{person}{Sebastian
  Herholz}, {and} \bibinfo{person}{Hendrik P.~A. Lensch}.}
  \bibinfo{year}{2020}\natexlab{}.
\newblock \showarticletitle{Robust fitting of parallax-aware mixtures for path
  guiding}.
\newblock \bibinfo{journal}{\emph{ACM Trans. Graph.}} \bibinfo{volume}{39},
  \bibinfo{number}{4}, Article \bibinfo{articleno}{147} (\bibinfo{date}{aug}
  \bibinfo{year}{2020}), \bibinfo{numpages}{15}~pages.
\newblock
\showISSN{0730-0301}
\urldef\tempurl%
\url{https://doi.org/10.1145/3386569.3392421}
\showDOI{\tempurl}


\bibitem[Simon et~al\mbox{.}(2018)]%
        {reibold2018selective}
\bibfield{author}{\bibinfo{person}{Florian Simon}, \bibinfo{person}{Alisa
  Jung}, \bibinfo{person}{Johannes Hanika}, {and} \bibinfo{person}{Carsten
  Dachsbacher}.} \bibinfo{year}{2018}\natexlab{}.
\newblock \showarticletitle{Selective guided sampling with complete light
  transport paths}.
\newblock \bibinfo{journal}{\emph{Transactions on Graphics (Proceedings of
  SIGGRAPH Asia)}} \bibinfo{volume}{37}, \bibinfo{number}{6}
  (\bibinfo{date}{Dec.} \bibinfo{year}{2018}).
\newblock
\urldef\tempurl%
\url{https://doi.org/10.1145/3272127.3275030}
\showDOI{\tempurl}


\bibitem[Snavely et~al\mbox{.}(2006)]%
        {snavely2006photo}
\bibfield{author}{\bibinfo{person}{Noah Snavely}, \bibinfo{person}{Steven~M
  Seitz}, {and} \bibinfo{person}{Richard Szeliski}.}
  \bibinfo{year}{2006}\natexlab{}.
\newblock \showarticletitle{Photo tourism: exploring photo collections in 3D}.
\newblock In \bibinfo{booktitle}{\emph{ACM siggraph 2006 papers}}.
  \bibinfo{pages}{835--846}.
\newblock


\bibitem[Sun et~al\mbox{.}(2022)]%
        {sun2022direct}
\bibfield{author}{\bibinfo{person}{Cheng Sun}, \bibinfo{person}{Min Sun}, {and}
  \bibinfo{person}{Hwann-Tzong Chen}.} \bibinfo{year}{2022}\natexlab{}.
\newblock \showarticletitle{Direct voxel grid optimization: Super-fast
  convergence for radiance fields reconstruction}. In
  \bibinfo{booktitle}{\emph{Proceedings of the IEEE/CVF Conference on Computer
  Vision and Pattern Recognition}}. \bibinfo{pages}{5459--5469}.
\newblock


\bibitem[Takikawa et~al\mbox{.}(2021)]%
        {takikawa2021neural}
\bibfield{author}{\bibinfo{person}{Towaki Takikawa}, \bibinfo{person}{Joey
  Litalien}, \bibinfo{person}{Kangxue Yin}, \bibinfo{person}{Karsten Kreis},
  \bibinfo{person}{Charles Loop}, \bibinfo{person}{Derek Nowrouzezahrai},
  \bibinfo{person}{Alec Jacobson}, \bibinfo{person}{Morgan McGuire}, {and}
  \bibinfo{person}{Sanja Fidler}.} \bibinfo{year}{2021}\natexlab{}.
\newblock \showarticletitle{Neural geometric level of detail: Real-time
  rendering with implicit 3d shapes}. In \bibinfo{booktitle}{\emph{Proceedings
  of the IEEE/CVF Conference on Computer Vision and Pattern Recognition}}.
  \bibinfo{pages}{11358--11367}.
\newblock


\bibitem[Tancik et~al\mbox{.}(2020)]%
        {tancik2020fourier}
\bibfield{author}{\bibinfo{person}{Matthew Tancik}, \bibinfo{person}{Pratul
  Srinivasan}, \bibinfo{person}{Ben Mildenhall}, \bibinfo{person}{Sara
  Fridovich-Keil}, \bibinfo{person}{Nithin Raghavan}, \bibinfo{person}{Utkarsh
  Singhal}, \bibinfo{person}{Ravi Ramamoorthi}, \bibinfo{person}{Jonathan
  Barron}, {and} \bibinfo{person}{Ren Ng}.} \bibinfo{year}{2020}\natexlab{}.
\newblock \showarticletitle{Fourier features let networks learn high frequency
  functions in low dimensional domains}.
\newblock \bibinfo{journal}{\emph{Advances in neural information processing
  systems}}  \bibinfo{volume}{33} (\bibinfo{year}{2020}),
  \bibinfo{pages}{7537--7547}.
\newblock


\bibitem[Tewari et~al\mbox{.}(2022)]%
        {tewari2022advances}
\bibfield{author}{\bibinfo{person}{Ayush Tewari}, \bibinfo{person}{Justus
  Thies}, \bibinfo{person}{Ben Mildenhall}, \bibinfo{person}{Pratul
  Srinivasan}, \bibinfo{person}{Edgar Tretschk}, \bibinfo{person}{Wang Yifan},
  \bibinfo{person}{Christoph Lassner}, \bibinfo{person}{Vincent Sitzmann},
  \bibinfo{person}{Ricardo Martin-Brualla}, \bibinfo{person}{Stephen Lombardi},
  {et~al\mbox{.}}} \bibinfo{year}{2022}\natexlab{}.
\newblock \showarticletitle{Advances in neural rendering}. In
  \bibinfo{booktitle}{\emph{Comp. Graph. Forum}}, Vol.~\bibinfo{volume}{41}.
  \bibinfo{pages}{703--735}.
\newblock


\bibitem[Verbin et~al\mbox{.}(2022)]%
        {verbin2022refnerf}
\bibfield{author}{\bibinfo{person}{Dor Verbin}, \bibinfo{person}{Peter Hedman},
  \bibinfo{person}{Ben Mildenhall}, \bibinfo{person}{Todd Zickler},
  \bibinfo{person}{Jonathan~T. Barron}, {and} \bibinfo{person}{Pratul~P.
  Srinivasan}.} \bibinfo{year}{2022}\natexlab{}.
\newblock \showarticletitle{{Ref-NeRF}: Structured View-Dependent Appearance
  for Neural Radiance Fields}.
\newblock \bibinfo{journal}{\emph{CVPR}} (\bibinfo{year}{2022}).
\newblock


\bibitem[Vorba et~al\mbox{.}(2014)]%
        {vorba2014line}
\bibfield{author}{\bibinfo{person}{Ji{\v{r}}{\'\i} Vorba},
  \bibinfo{person}{Ond{\v{r}}ej Karl{\'\i}k}, \bibinfo{person}{Martin
  {\v{S}}ik}, \bibinfo{person}{Tobias Ritschel}, {and}
  \bibinfo{person}{Jaroslav K{\v{r}}iv{\'a}nek}.}
  \bibinfo{year}{2014}\natexlab{}.
\newblock \showarticletitle{On-line learning of parametric mixture models for
  light transport simulation}.
\newblock \bibinfo{journal}{\emph{ACM Trans. Graph.}} \bibinfo{volume}{33},
  \bibinfo{number}{4} (\bibinfo{year}{2014}), \bibinfo{pages}{1--11}.
\newblock


\bibitem[Wang et~al\mbox{.}(2021)]%
        {wang2021neus}
\bibfield{author}{\bibinfo{person}{Peng Wang}, \bibinfo{person}{Lingjie Liu},
  \bibinfo{person}{Yuan Liu}, \bibinfo{person}{Christian Theobalt},
  \bibinfo{person}{Taku Komura}, {and} \bibinfo{person}{Wenping Wang}.}
  \bibinfo{year}{2021}\natexlab{}.
\newblock \showarticletitle{NeuS: Learning Neural Implicit Surfaces by Volume
  Rendering for Multi-view Reconstruction}.
\newblock \bibinfo{journal}{\emph{Advances in Neural Information Processing
  Systems}}  \bibinfo{volume}{34} (\bibinfo{year}{2021}),
  \bibinfo{pages}{27171--27183}.
\newblock


\bibitem[Wizadwongsa et~al\mbox{.}(2021)]%
        {Wizadwongsa2021NeX}
\bibfield{author}{\bibinfo{person}{Suttisak Wizadwongsa},
  \bibinfo{person}{Pakkapon Phongthawee}, \bibinfo{person}{Jiraphon
  Yenphraphai}, {and} \bibinfo{person}{Supasorn Suwajanakorn}.}
  \bibinfo{year}{2021}\natexlab{}.
\newblock \showarticletitle{NeX: Real-time View Synthesis with Neural Basis
  Expansion}. In \bibinfo{booktitle}{\emph{IEEE Conference on Computer Vision
  and Pattern Recognition (CVPR)}}.
\newblock


\bibitem[Yu et~al\mbox{.}(2021)]%
        {yu2021plenoctrees}
\bibfield{author}{\bibinfo{person}{Alex Yu}, \bibinfo{person}{Ruilong Li},
  \bibinfo{person}{Matthew Tancik}, \bibinfo{person}{Hao Li},
  \bibinfo{person}{Ren Ng}, {and} \bibinfo{person}{Angjoo Kanazawa}.}
  \bibinfo{year}{2021}\natexlab{}.
\newblock \showarticletitle{Plenoctrees for real-time rendering of neural
  radiance fields}. In \bibinfo{booktitle}{\emph{Proceedings of the IEEE/CVF
  International Conference on Computer Vision}}. \bibinfo{pages}{5752--5761}.
\newblock


\bibitem[Zhao et~al\mbox{.}(2020)]%
        {zhao2020physics}
\bibfield{author}{\bibinfo{person}{Shuang Zhao}, \bibinfo{person}{Wenzel
  Jakob}, {and} \bibinfo{person}{Tzu-Mao Li}.} \bibinfo{year}{2020}\natexlab{}.
\newblock \showarticletitle{Physics-based differentiable rendering: a
  comprehensive introduction}.
\newblock \bibinfo{journal}{\emph{ACM SIGGRAPH 2020 Courses}}
  \bibinfo{volume}{14} (\bibinfo{year}{2020}), \bibinfo{pages}{1--14}.
\newblock


\bibitem[Zhou et~al\mbox{.}(2008)]%
        {zhou2008real}
\bibfield{author}{\bibinfo{person}{Kun Zhou}, \bibinfo{person}{Zhong Ren},
  \bibinfo{person}{Stephen Lin}, \bibinfo{person}{Hujun Bao},
  \bibinfo{person}{Baining Guo}, {and} \bibinfo{person}{Heung-Yeung Shum}.}
  \bibinfo{year}{2008}\natexlab{}.
\newblock \showarticletitle{Real-time smoke rendering using compensated ray
  marching}.
\newblock In \bibinfo{booktitle}{\emph{ACM Trans. Graph. (SIGGRAPH)}}.
  \bibinfo{pages}{1--12}.
\newblock


\end{thebibliography}

\end{document}